\documentclass{article}
\usepackage{spconf,amsmath,epsfig}
\usepackage{amssymb}
\usepackage{algorithm}
\usepackage{algorithmic}

\usepackage{multirow}
\usepackage{multicol}
\usepackage{color}

\usepackage{hyperref}
\hypersetup{hidelinks,
	colorlinks=true,
	allcolors=black,
	pdfstartview=Fit,
	breaklinks=true}

\let\OLDthebibliography\thebibliography
\renewcommand\thebibliography[1]{
  \OLDthebibliography{#1}
  \setlength{\parskip}{0pt}
  \setlength{\itemsep}{0pt plus 0.3ex}
}

\begin{document}\sloppy

\def\x{{\mathbf x}}
\def\L{{\cal L}}

\title{Collaborative Auto-encoding for Blind Image Quality Assessment}
%
\name{\textit{Zehong Zhou}$^{1,3,4}$, \textit{Fei Zhou}$^{1,2,3,6}$\dag, \textit{Guoping Qiu}$^{1,2,3,4,5,6}$}
\address{$^{1}$College of Electronic and Information Engineering, Shenzhen University, Shenzhen, China \\
$^{2}$Peng Cheng Laboratory, Shenzhen, China \\
$^{3}$Guangdong Key Laboratory of Intelligent Information Processing, Shenzhen, China \\
$^{4}$Shenzhen Institute for Artificial Intelligence and Robotics for Society, Shenzhen, China \\
$^{5}$School of Computer Science, University of Nottingham, Nottingham, U.K \\
$^{6}$Guangdong-Hong Kong Joint Laboratory for Big Data Imaging and Communication, Shenzhen, China \\
Email: flying.zhou@163.com
}

\maketitle

\begin{abstract}
Blind image quality assessment (BIQA) is a challenging problem with important real-world applications. 
Recent efforts attempting to exploit powerful representations by deep neural networks (DNN) are hindered by the lack of subjectively annotated data. 
This paper presents a novel BIQA method which overcomes this fundamental obstacle. 
Specifically, we design a pair of collaborative autoencoders (COAE) consisting of a content autoencoder (CAE) and a distortion autoencoder (DAE) that work together to extract content and distortion representations, which are shown to be highly descriptive of image quality. 
While the CAE follows a standard codec procedure, we introduce the CAE-encoded feature as an extra input to the DAE’s decoder for reconstructing distorted images, thus effectively forcing DAE’s encoder to extract distortion representations. 
The self-supervised learning framework allows the COAE including two feature extractors to be trained by almost unlimited amount of data, thus leaving limited samples with annotations to finetune a BIQA model. 
We will show that the proposed BIQA method achieves state-of-the-art performance and has superior generalization capability over other learning based models. The codes are available at: \href{https://github.com/Macro-Zhou/NRIQA-VISOR}{https://github.com/Macro-Zhou/NRIQA-VISOR/}. 
\end{abstract}
\section{Introduction}
\label{sec:intro}

Image quality assessment (IQA) is an indispensable tool in the modern digital media era. IQA models can be divided as full reference (FR), reduced reference (RR), and no reference (NR) according to whether full, partial, and no reference image is available. Compared with FR IQA and RR IQA, NR IQA, also known as blind image quality assessment (BIQA), is much more challenging and potentially more useful.

In BIQA, the key is to extract visual quality-related features.  Many traditional methods \cite{brisque2012, nferm2015, ilniqe2015} extract such features based on natural scene statistics (NSS). However, due to the complexity of distortion and diversity of image content, these hand-crafted features have limited visual quality representation capabilities. 
Recently, deep neural networks (DNNs) have been employed in BIQA. However, the problem of insufficient annotated data, e.g., not enough images with mean opinion scores (MOS), is hindering the development of deep learning-based BIQA solutions. To address the problem, one approach is to develop end-to-end models using various learning strategies including patch-wise learning \cite{wadiqam2018}, 
multi-task learning \cite{meon2018}, 
transfer learning \cite{dbcnn2020}, meta-learning \cite{metaplus2021}, and so on. Despite progress, most existing DNN-based BIQA methods generally suffer from the “blackbox” problem where it is often unknown what features are extracted and what role a feature plays. Another solution is to use supervised learning to extract relevant features through: (i) manually designing different learning targets such as pseudo MOS / similarity maps \cite{diqa2019} and ranking labels \cite{clriqa2022}, or (ii) performing quality-related tasks such as image restoration \cite{vcrnet2022} and primary content prediction \cite{aigqa2021}. However, their effectiveness heavily depends on the designers' experience. Therefore, the problem of lacking annotated data remains a fundamental obstacle to developing deep learning-based BIQA. 

In this paper, we tackle these issues by designing an autoencoder (AE) system through self-supervised learning, thus overcoming the difficulty of having to annotate large amount of training data. Using self-supervised learning means that there is almost unlimited amount of data available for training highly representative models to extract 
quality-related features, thus leaving the limited MOS annotated samples to train a light-weight quality predictor.
The core of our novel BIQA method is the introduction of a COllaborative AutoEncoder (COAE) to extract image content and distortion representations. The COAE consists of a content autoencoder (CAE) and a distortion autoencoder (DAE) working collaboratively and individually. While the CAE is a standard AE trained with high quality undistorted images for extracting content-related representations, the design of DAE is unique and key to the success of the new BIQA model. By introducing the CAE-encoded content feature as an external input to the decoder of the DAE and requiring the DAE’s decoder to reconstruct distorted input images, the DAE’s encoder is forced to extract distortion-aware features. As image quality is not only related to distortion but is also content dependent, the content features from the CAE and the distortion features from the DAE are together sent to a quality score regressor to assess the input image's quality without using a reference. 

\begin{figure*}[t]
    \centering
    \includegraphics[width=\textwidth]{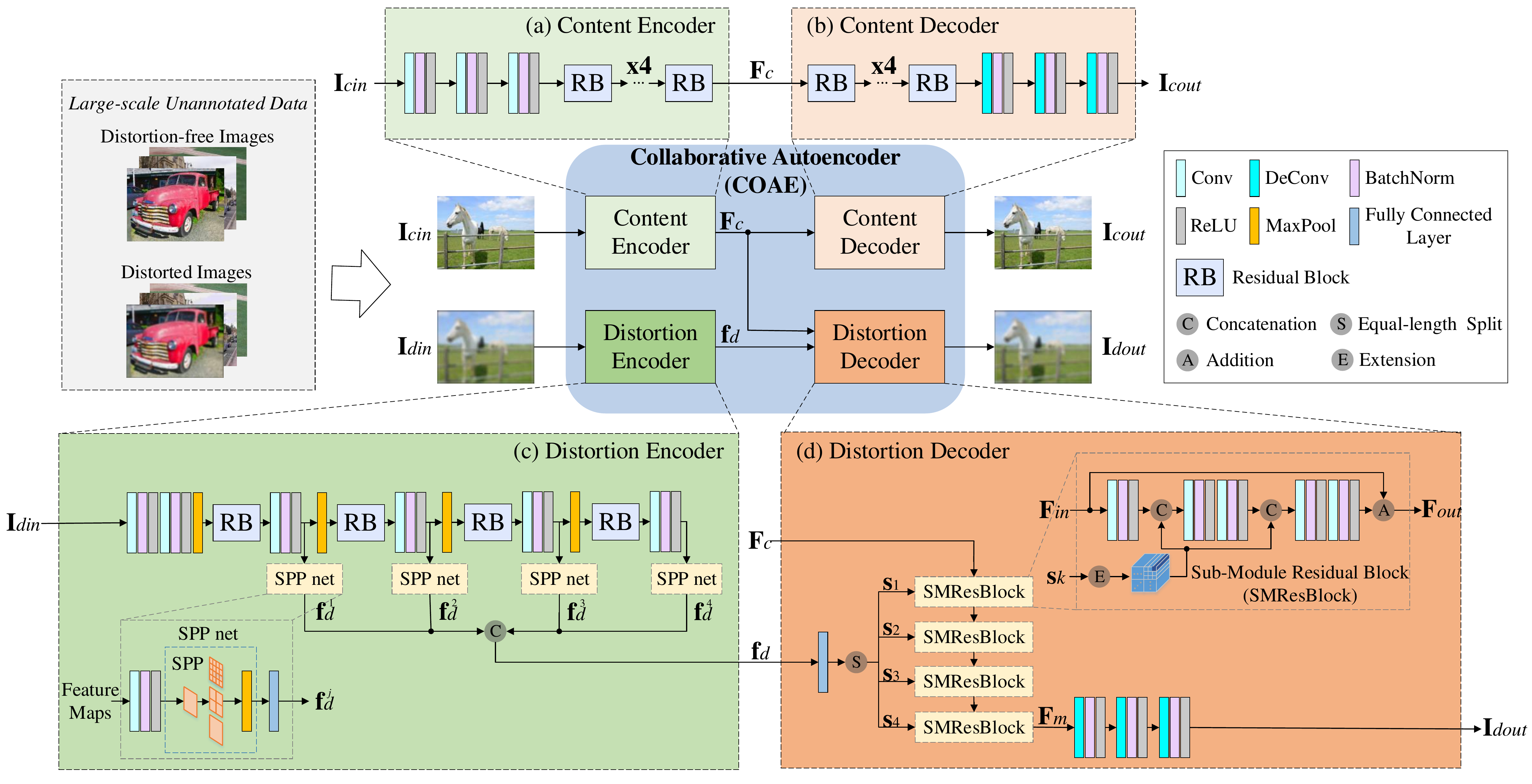}
    \caption{The proposed collaborative autoencoder.
}
    \label{fig1:COAE}
\end{figure*}

The main contributions of this paper can be summarized as follows:

(1) We propose a self-supervised learning strategy to extract quality-related features for developing deep learning-based BIQA models, thus overcoming the fundamental problem of lacking sufficient annotated data.

(2) We introduce an innovative collaborative autoencoder (COAE) architecture and design a strategy to enable a pair of autoencoders to work collaboratively and individually to extract content feature and distortion-aware feature respectively. We will show that these features are highly descriptive of image quality. 

(3) We have developed a BIQA method termed Visual qualIty aSsessment based on cOllaborative autoencodeRs (VISOR) which uses the content feature and the distortion-aware feature of the COAE to construct a blind image quality predictor. We will show that VISOR achieves state-of-the-art performance and has superior generalization capability over other learning based models.

\section{Collaborative Autoencoder}

Image quality mainly depends on two attributes: the image content and distortions \cite{clearskyandblur2019}. 
However, when taking image as a signal, the content information mostly dominants the signal energy, while the distortion information is much smaller and regarded as disturbance. 
Therefore, when training a single model in a normal self-supervised way (e.g., an AE), it tends to encode the main content information even if it is trained on distorted images. 
When using such features for quality assessment, the distortion information is usually difficult to make enough contributions. 
To settle that, a novel self-supervised architecture, termed as collaborative autoencoder (COAE), is introduced to separately extract content- and distortion-related features, as shown in Fig. \ref{fig1:COAE}. 
The new architecture consists of two AEs, the CAE and the DAE. 
The CAE is trained as a standard AE using pristine images, aiming at extracting main information of an image, i.e., the content features. 
However, when training a DAE, our idea is to externally injecting its decoder with sufficient content information to force its encoder to concentrate on image distortions. 
If trained on distorted images, the mission of the decoder is to better reconstruct the distorted input images. With the content information from CAE, the DAE's encoder only needs to focus on the distortion information to enable the DAE's decoder to make accurate reconstruction. In this way, the DAE can effectively learn the distortion representation. 
In the following subsections, we detail the architecture of COAE.

\subsection{The Content Autoencoder}
It has been shown in many vision tasks that the AEs can extract content features of the input images in a self-supervised way. Thus, it is reasonable to use a standard AE to extract main content information. Specifically, the CAE consists of an encoder $\mathbb{E}_c$ employed to extract content features $\textbf{F}_c$ from the input image $\textbf{I}_{cin}$, and a decoder $\mathbb{D}_c$ used to reproduce the input image from $\textbf{F}_c$. 
As shown in the Fig. \ref{fig1:COAE}(a) and (b), 
the architecture of $\mathbb{E}_c$ has three convolutional layers with the appliance of batch normalization and ReLU activation, which are subsequently followed by four residual blocks.  
The architecture of $\mathbb{D}_c$ is dual with $\mathbb{E}_c$: four residual blocks are followed by three deconvolutional layers. 
The feature channel of $\textbf{F}_c$ is set to 256 since the image contents are diversed.

To ensure $\textbf{F}_c$ will not be disturbed by distortions, only pristine images are utilized to train the CAE using the following loss function: 
\begin{equation}
\label{eq_CAE_loss}
    L_{overall}= L_{recon} + L_{percep},
\end{equation}
where $L_{recon}=\left\|\textbf{I}_{cin}-\textbf{I}_{cout}\right\|_F$ is the pixel-wise loss, $\left\|.\right\|_F$ is the Frobenius norm, and $L_{percep}=LPIPS(\textbf{I}_{in}, \textbf{I}_{out})$ is the Learned Perceptual Image Patch Similarity \cite{lpips2018} between $\textbf{I}_{in}$ and $\textbf{I}_{out}$ representing the perceptual loss, which is commonly used in reconstruction tasks \cite{aigqa2021}. 

\subsection{The Distortion Autoencoder}
Different from the CAE, the DAE's decoder has an extra input, i.e., the content features $\textbf{F}_c$ from the CAE's encoder. When the input images of the CAE and the DAE share the same image content, the $\textbf{F}_c$ will contain adequate content information for DAE's decoder to reconstruct DAE's input image. By this way, when the output of the DAE's encoder is restricted to a low dimension, the encoder will strive to abandon most of the content information and extract distortion information to enable the decoder to make accurate reconstruction. 
Specifically, a distortion encoder $\mathbb{E}_d$ is employed to extract the distortion-aware features $\textbf{f}_d$ from the input $\textbf{I}_{din}$. Using both $\textbf{f}_d$ and $\textbf{F}_c$, a distortion decoder $\mathbb{D}_d$ is designed to reproduce $\textbf{I}_{din}$. 
The procedure of the DAE can be described as:
\begin{equation}
    \textbf{I}_{dout}=\mathbb{D}_d(\textbf{F}_c,\textbf{f}_d)=\mathbb{D}_d(\mathbb{E}_c(\textbf{I}_{cin}),\mathbb{E}_d(\textbf{I}_{din})),
\end{equation}
where $\textbf{I}_{din}$ is a distorted version of $\textbf{I}_{cin}$ and  $\textbf{I}_{dout}$ is the output of the DAE.

The architecture of the distortion encoder is illustrated in Fig. \ref{fig1:COAE}(c).
As distortions may affect local, regional and global structures, features extracted from multiple layers are included. 
Specifically, four feature maps are extracted from four different layers. 
To further learn multi-scale information, each map is fed to a Spatial Pyramid Pooling (SPP) net to generate a fixed-length representation. The SPP net is a simple network consisting of a convolutional layer, an SPP operation \cite{spp2015}, and an FC layer. 
After obtaining $\textbf{f}_d^j$ from multiple layers, they are concatenated to form the distortion-aware feature: $\textbf{f}_d=concat(\textbf{f}_d^1,\textbf{f}_d^2,\textbf{f}_d^3,\textbf{f}_d^4)\in\mathcal{R}^{256\times1}$.
The distortion decoder is shown in Fig. \ref{fig1:COAE}(d).  
It has two inputs, the content feature $\textbf{F}_c$ and the distortion-aware feature $\textbf{f}_d$. 
Using an effective technique similar to feature modulation \cite{disentangled2018}, we treat $\textbf{f}_d$ as a modulation signal to rectify $\textbf{F}_c$. Specifically, $\textbf{f}_d$ is first fed into an FC layer to generate an extended feature $\textbf{s}\in\mathcal{R}^{1024\times1}$, 
which is then split evenly into four features \{$\textbf{s}_k\in\mathcal{R}^{256\times1}, k=1,2,3,4$\} before feeding into four Sub-Modulation Residual Blocks (SMResBlocks) to modulate $\textbf{F}_c$ in sequence. In each SMResBlock, $\textbf{s}_k$ is spatially extended into feature maps with the same width and height as $\textbf{F}_c$ through simple copying.  
After being modulated four times with four SMResBlocks, the modulated feature $\textbf{F}_m$ is obtained from $\textbf{F}_c$ and $\textbf{f}_d$. After feature modulation, we further employ three deconvolutional layers with the same kernel size and stride as those in the content decoder $\mathbb{D}_c$ to transform $\textbf{F}_m$ into $\textbf{I}_{dout}$. The same loss function $L_{overall}$ as described in Eq. (\ref{eq_CAE_loss}) is adopted to optimize the DAE's reconstruction where $\textbf{I}_{cin}$ and $\textbf{I}_{cout}$ are respectively replaced by $\textbf{I}_{din}$ and $\textbf{I}_{dout}$.

\begin{figure}[t]
    \centering
    \includegraphics[width=\linewidth]{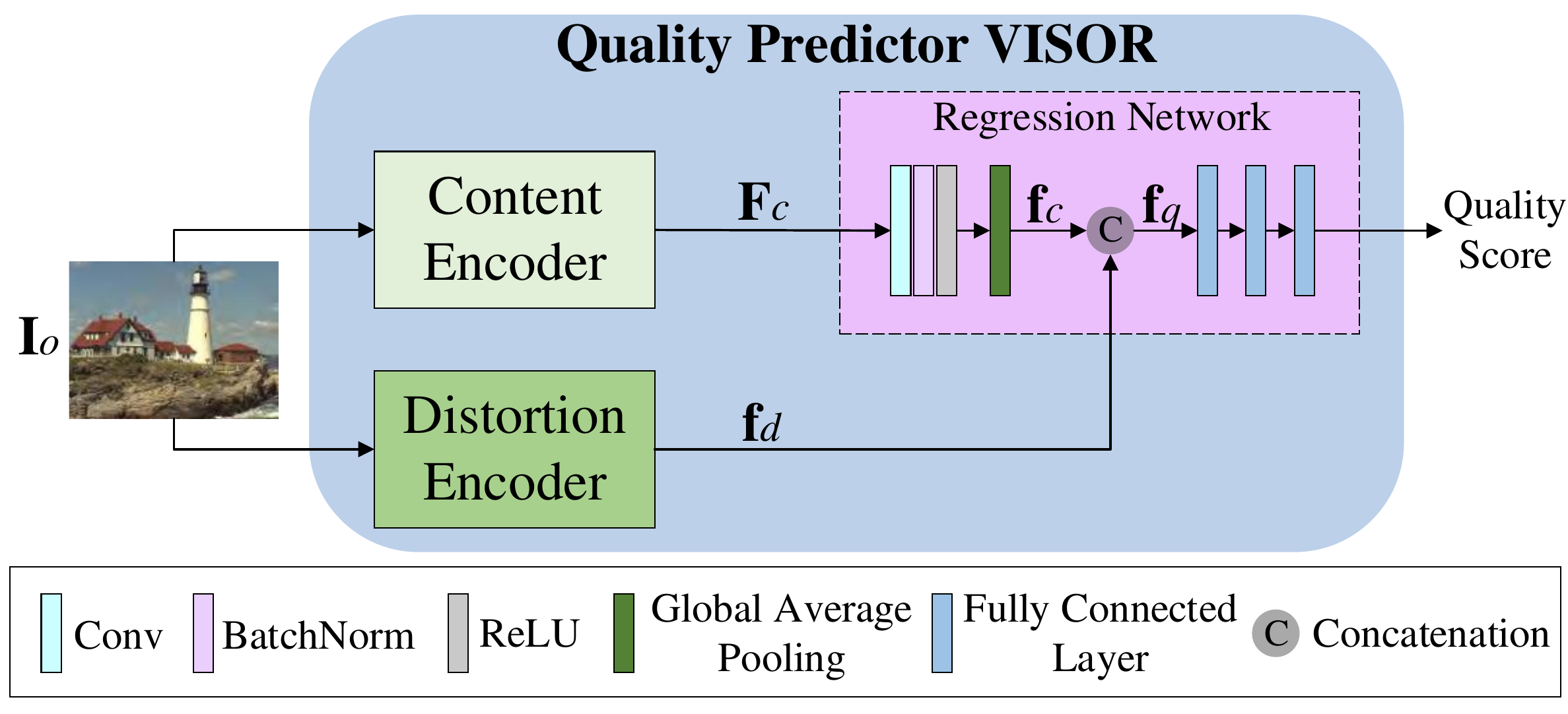}
    \caption{The proposed BIQA model VISOR.
}
    \label{fig2:VISOR}
\end{figure}

\subsection{Training Procedure of the COAE}
The training of the COAE has two stages. In the first stage, the CAE is trained with a large number of pristine images to provide pure $\textbf{F}_c$. In the second stage, the parameters of the CAE are frozen so that the $\textbf{F}_c$ can be stable. Then a number of distorted images are utilized to train the DAE, when CAE's inputs are still pristine images that share the same content with the distorted ones. After all the learnable parameters in the COAE are determined, we obtain two feature extractors: the content encoder $\mathbb{E}_c$ and the distortion encoder $\mathbb{E}_d$.  Several experiments will be conducted to demonstrate the effectiveness of these two feature extractors.

\begin{table*}
    \caption{Performance comparisons on individual IQA datasets.}
    \label{tab2:my_label}
    \centering
    \resizebox{\linewidth}{!}{
    \begin{tabular}{l|cc|cc|cc|cc|cc|cc|cc}
        \cline{1-15}
        \multirow{2}{*}{Method} & \multicolumn{2}{|c|}{LIVE} & \multicolumn{2}{|c|}{CSIQ} & \multicolumn{2}{|c|}{TID2013} & \multicolumn{2}{|c|}{KADID-10k} & \multicolumn{2}{|c|}{LIVEC} & \multicolumn{2}{|c}{KonIQ-10k} & \multicolumn{2}{|c}{Weigthed Avg.} \\ \cline{2-15}
         & SRCC & PLCC & SRCC & PLCC & SRCC & PLCC & SRCC & PLCC & SRCC & PLCC & SRCC & PLCC & SRCC & PLCC \\ \cline{1-15}
        BRISQUE & 0.939 & 0.942 & 0.750 & 0.829 & 0.573 & 0.651 & 0.422 & 0.491 & 0.607 & 0.585 & 0.624 & 0.624 & 0.552 & 0.590 \\
        NFERM & 0.941 &	0.946 &	0.769 &	0.811 &	0.584 &	0.660 & 0.502 &	0.583 &	0.536 &	0.581 &	0.590 &	0.656 & 0.569 & 0.638 \\
        ILNIQE & 0.902 & 0.865 & 0.807 & 0.808 & 0.519 & 0.640 & 0.533 & 0.571 & 0.430 & 0.510 & 0.507 & 0.523 & 0.537 & 0.574 \\
        HOSA & 0.948 &	0.949 &	0.781 &	0.842 &	0.688 &	0.764 & 0.613 & 0.651 & 0.660 & 0.680 & 0.805 & 0.813 & 0.714 & 0.743 \\ \cline{1-15}
        WaDIQaM & 0.960 & 0.972 & 0.852 & 0.844 & 0.835 & 0.855 & \underline{0.885} & \underline{0.890} & 0.606 & 0.601 & 0.710 & 0.738 & 0.800 & 0.815\\ 
        DIQA & \underline{0.975} & 0.977 & 0.884 & 0.915 & 0.825 & 0.850 & / & / & 0.703 & 0.704 & / & / & / & / \\
        DBCNN & 0.968 & 0.971 & 0.937 & 0.938 & 0.816 & 0.865 & 0.851 & 0.856 & 0.844 & 0.851 &	0.864 & 0.872 & 0.858 & 0.869 \\ 
        AIGQA & 0.960 & 0.957 & 0.927 & 0.952 & \underline{0.871} & \underline{0.893} & 0.864 & 0.863 & 0.751 & 0.761 & / & / & / & / \\
        VCRNet & 0.973 & 0.974 & \underline{0.945} & \underline{0.955} & 0.846 & 0.875 & 0.853 & 0.849 & \textbf{0.856} & \underline{0.865} & \underline{0.894} & \underline{0.909} & \underline{0.866} & \underline{0.876}\\ 
        CLRIQA & \textbf{0.977} & \textbf{0.980} & 0.915 & 0.938 & 0.837 & 0.863 & 0.837 & 0.843 & 0.832 & \textbf{0.866} & 0.831 & 0.846 & 0.841 & 0.855 \\ \cline{1-15}
        VISOR & 0.973 & \underline{0.978} & \textbf{0.961} & \textbf{0.967} & \textbf{0.905} & \textbf{0.922} & \textbf{0.928} & \textbf{0.930} & \underline{0.846} & \textbf{0.866} & \textbf{0.896} & \textbf{0.910} & \textbf{0.911} & \textbf{0.921} \\
        \cline{1-15}
    \end{tabular}
    }
\end{table*}

\section{The COAE based BIQA Model}

Since the image quality mainly depends on two attributes: the primary content and distortions \cite{dbcnn2020, aigqa2021, imagewise2017spm}, both $\mathbb{E}_c$ and $\mathbb{E}_d$ are utilized to build an effective BIQA model. The overview of the COAE based BIQA model VISOR is shown in Fig. \ref{fig2:VISOR}. 
Since the CAE is trained on pristine images only, $\textbf{F}_c$ from the distorted input images will be contaminated by distortion. To reduce the effect of the distortion on $\textbf{F}_c$, we first reduce the dimension of $\textbf{F}_c$ to output a content-aware feature $\textbf{f}_c\in\mathcal{R}^{16\times1}$ by feeding it into a simple pooling module consisting of one convolutional layer with a 1×1 kernel (for channel-wise pooling) and one global average pooling (GAP) layer. 
Then $\textbf{f}_c$ and $\textbf{f}_d$ are concatenated as the input $\textbf{f}_q\in\mathcal{R}^{272\times1}$ to a regression network consisting of three FC layers to simulate the interaction of the two attributes and infer the quality score of the input image. 

When training the VISOR, the parameters of $\mathbb{E}_c$ and $\mathbb{E}_d$ are \textbf{FROZEN}. Therefore, only few parameters need to be learned in the VISOR, which can be fully trained on existing IQA datasets with subjective annotations. The VISOR is optimized by minimizing the mean square error (MSE) between the predicted scores and the ground-truth labels. 

\section{Experiments}

\subsection{Implementation Details and Data
}

\textbf{Implementation.} The proposed COAE and VISOR is implemented in Pytorch on a machine of 4 NVIDIA Tesla P40 GPUs. When training the COAE, all the training images are cropped into 256×256 patches before fed into the network. 
When training the VISOR, the images as well as their subjective scores in existing IQA datasets are utilized, and the parameters of $\mathbb{E}_c$ and $\mathbb{E}_d$ are frozen. The 224×224 patches are cropped from the input images to fit the mini-batch training, while the full-size images can be directly used as the inputs in the testing phases owing to the introduction of SPP module. 

\textbf{Data Preparation for COAE.} To fully train the COAE, we have prepared a large amount of data. 
Specifically, we collect and pick out 10, 000 high-quality images from ImageNet \footnote{https://image-net.org/} and MSCOCO \footnote{https://cocodataset.org/} as pristine images. Afterwards, these images are destructed with 25 distortion types at 5 levels, which cover most distortions that might occur in the real world. 
Totally, 10, 000 pristine images and 1, 250, 000 distorted images are employed to train the COAE. 

\textbf{IQA Datasets and Criteria.} Four synthetic IQA datasets including LIVE \cite{live2006}, CSIQ \cite{csiq2010}, TID2013 \cite{tid2013}, and KADID-10K \cite{kadid2020}, and two authentic IQA datasets including LIVEC \cite{livec2016} and KonIQ-10K \cite{koniq10k2020}, are used to evaluate the performance of the proposed VISOR method. To evaluate the performance of the BIQA models, two commonly-used criteria, the Spearman Rank Order Correlation Coefficient (SRCC) and the Pearson Linear Correlation Coefficient (PLCC) are employed to evaluate the monotonicity and linearity between the predictions and the labels. 

\subsection{Performance Comparisons}
We test the performance of VISOR on six IQA datasets and compare it with 10 classical and state-of-the-art BIQA methods. The competitors include 4 traditional BIQA methods (BRISQUE \cite{brisque2012}, NFERM \cite{nferm2015}, ILNIQE \cite{ilniqe2015}, and HOSA \cite{hosa2015}) and 6 DNN-based BIQA methods (WaDIQaM \cite{wadiqam2018}, DIQA \cite{diqa2019}, DBCNN \cite{dbcnn2020}, AIGQA \cite{aigqa2021}, CLRIQA \cite{clriqa2022}, and VCRNet \cite{vcrnet2022}). 
When evaluating on each dataset, all the images are randomly divided into 80\% for training and 20\% for testing \cite{imagewise2017spm}. For the synthetic datasets, each dataset is divided according to the reference image to ensure that there is no overlapping image content between the training and testing sets. The experiments are conducted in 10 sessions and the median SRCC, PLCC results are reported to avoid the training bias.

Table \ref{tab2:my_label} lists the SRCC and PLCC results on different datasets. The best and second-best results of SRCC and PLCC are respectively highlighted in bold and underline. As can be seen, the proposed VISOR achieves the bes or second-best results in 11 of 12 comparisons across six IQA datasets. 
Note that the VISOR achieves three best results and one second-best result on two authentic datasets LIVEC and KonIQ-10k, even though its feature extractors were only trained with synthetic data.
We attribute this to the implementation of several distortion types that simulate real-world distortions (such as motion blur, under- and over-exposure, etc.) when training the COAE, and the separate representations of content and distortion information. 
The results on six datasets and the overall weighted average performance indicate that the VISOR shows its superiority over the other methods. 

\begin{table}
    \caption{Cross-Dataset Evaluation}
    \label{tab4:my_label}
    \centering
    \begin{tabular}{l|c|c|c|c}
        \cline{1-5}
        Train & \multicolumn{2}{|c|}{TID2013} &  \multicolumn{2}{|c}{KonIQ-10k} \\ \cline{1-5}
        Test & \multicolumn{1}{|c|}{LIVE} & \multicolumn{1}{|c}{CSIQ} 
        & \multicolumn{1}{|c}{LIVE} & \multicolumn{1}{|c}{LIVEC}\\ \cline{1-5}
        BRISQUE \cite{brisque2012} & 0.814 & 0.612 & 0.336 & 0.525 \\
        NFERM \cite{nferm2015} & 0.634 & 0.551 & 0.502 & 0.493\\
        WaDIQaM \cite{wadiqam2018} & 0.792 & 0.690 & 0.277 & 0.646 \\
        DBCNN \cite{dbcnn2020} & 0.891 & 0.807 & 0.676 & 0.749 \\
        VCRNet \cite{vcrnet2022} & 0.822 & 0.712 & 0.401 & 0.616 \\ \cline{1-5}
        VISOR & \textbf{0.909} & \textbf{0.870} & \textbf{0.715} & \textbf{0.754} \\
        \cline{1-5}
    \end{tabular}
\end{table}

\subsection{Cross-dataset Evaluations}
To analyze the generalization ability of the VISOR, cross-dataset evaluations are conducted. All the compared methods 
are trained on one specific dataset and tested on different datasets. The SRCC and PLCC results are listed in Table \ref{tab4:my_label}. As we can see, from the overall performance of both synthetic and authentic datasets, the generalization ability of VISOR outperforms others. 

\subsection{Ablation Studies}
Here we conduct ablation studies of the VISOR on TID2013 and LIVEC. 
First, 
to demonstrate the importance of the collaborative design of the COAE framework, 
we train the CAE and the DAE independently instead of collaboratively (denoted as \textbf{S-CAE} and \textbf{S-DAE}) with only distorted images. By this case, both of them can only extract mixed features.
We build 3 BIQA models using the S-CAE's feature, the S-DAE's feature, and the concatenated S-CAE and S-DAE feature (denoted as \textbf{S-CAE(C)S-DAE}), respectively, by sending them to a regression network.  
The performance is listed at the top part of the Table \ref{tab5:my_label}. It can be seen that without the collaboration of the two autoencoders, the performance of the extracted features drop dramatically, indicating the importance of the innovative design of the COAE framework.

\begin{table}
    \caption{IQA Performance with different settings}
    \label{tab5:my_label}
    \centering
    \begin{tabular}{c|cc|cc}
        \cline{1-5}
        \multirow{2}{*}{Setting} & \multicolumn{2}{|c|}{TID2013} & \multicolumn{2}{|c}{LIVEC} \\ \cline{2-5}
        & SRCC & PLCC & SRCC & PLCC \\ \cline{1-5}
        S-CAE & 0.766 & 0.799 & 0.719 & 0.745 \\
        S-DAE & 0.709 & 0.729 & 0.736 & 0.762 \\
        S-CAE(C)S-DAE & 0.769 & 0.801 & 0.740 & 0.765 \\ \cline{1-5}
        w/o SPP in DAE & 0.787 & 0.812 & 0.713 & 0.689 \\ 
        w/o ML in DAE & 0.755 & 0.777 & 0.720 & 0.740 \\ 
        Full DAE & 0.890 & 0.896 & 0.771 & 0.781 \\ \cline{1-5}
        CAE(C)DAE & \textbf{0.905} & \textbf{0.922} & \textbf{0.846} & \textbf{0.866} \\
        \cline{1-5}
    \end{tabular}
\end{table}

Second, to demonstrate the importance of the SPP nets and the multi-level feature extraction in DAE for efficiently capturing distortion information, we remove all the SPP nets in $\mathbb{E}_d$ and directly concatenate the distortion feature maps with the content features together inside $\mathbb{D}_d$ when training the COAE, and using the GAP results of the feature maps when training the VISOR. Then we conduct another study by only leaving the last layer output of the $\mathbb{E}_d$ instead of multi-level extraction. 
The results are listed at the middle part of the Table \ref{tab5:my_label}, which shows the necessity of the SPP nets and multi-level features extraction. 
Besides, the introduction of the content features further improves the performance, especially on authentic dataset LIVEC where the image contents are diverse, indicating the importance of the content information to quality assessment.

\begin{figure}
    \centering
    \includegraphics[width=\linewidth]{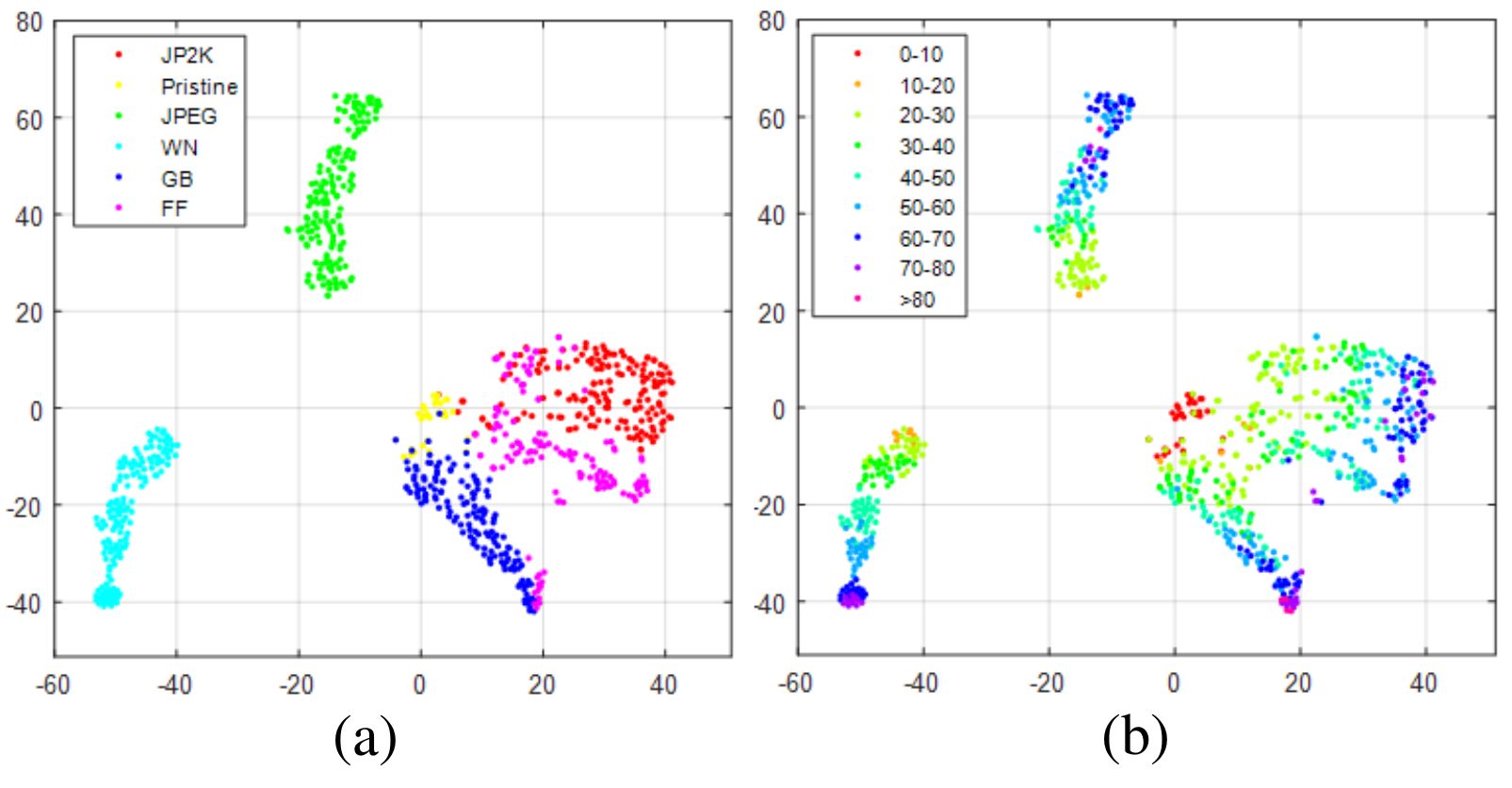}
    \caption{t-SNE results of the $\textbf{f}_d$. (a)
Scatter points colorized according to the distortion types of
images. (b) Scatter points colorized according to the DMOS
of images. 
}
    \label{fig_tsne}
\end{figure}

\subsection{Feature Analysis}
\textbf{Distortion-aware Features:} To analyse the effectiveness of distortion-aware features, we use t-SNE \cite{tsne2008} to visualize them. Specifically, $\textbf{f}_d$ are extracted from the distorted and pristine images in LIVE dataset, and then t-SNE algorithm is utilized to map the features onto a 2D visualization space as shown in Fig. \ref{fig_tsne}. Fig. \ref{fig_tsne}(a) shows the features for different distortions. It is seen that almost all the images are grouped together according to their distortion types. 
Fig. \ref{fig_tsne}(b) shows the features according to the images' DMOS. It is interesting to observe that pristine images are grouped together in the region near the origin. Besides, images with lower DMOS (higher quality) are closer to this area, whereas those with higher DMOS (lower quality) are further away from the origin. From the above analysis we can conclude that the distortion-aware features extracted by the novel COAE framework can effectively represent not only the characteristics of distortions but also the visual quality of the images. 

\textbf{Content-aware Features:} The content feature extracted by the CAE's encoder which is trained by pristine images is processed by a simple convolution and a GAP operations as shown in Fig. \ref{fig2:VISOR} to force it to discard most of the distortion information. To demonstrate the effectiveness of this scheme, the averaged cosine similarity between the $\textbf{F}_c$ of distorted images and their pristine versions for over one million samples in our COAE training data is calculated and the result is 0.974. The same calculation is also conducted on the LIVE, CSIQ, TID2013, and KADID-10k datasets, where the pristine versions of distorted images are available, and the averaged result is 0.964. The above results show that the content-aware features can indeed effectively represent the main information of the image content and are insensitive to distortions.

\section{Concluding Remarks}
In this work, we first propose a novel self-supervised framework based on a collaborative autoencoder (COAE) to extract effective quality-related features. 
Through employing a large-scale unlabeled data to fully train the COAE, effective content encoder and distortion encoder are obtained and then deployed to develop a BIQA model. Extensive experiments show the effectiveness of the proposed feature extraction framework 
and the excellent performance of the new blind image quality assessment model.


\end{document}